\documentclass[conference]{IEEEtran}
\IEEEoverridecommandlockouts

\usepackage{amsmath,amssymb,amsfonts}
\usepackage{booktabs}
\DeclareMathOperator*{\argmax}{arg\,max}
\usepackage{algorithm}
\usepackage{algorithmic}
\usepackage{makecell}
\usepackage{graphicx}
\usepackage{textcomp}
\usepackage{bbm}
\usepackage{xcolor}
\def\BibTeX{{\rm B\kern-.05em{\sc i\kern-.025em b}\kern-.08em
    T\kern-.1667em\lower.7ex\hbox{E}\kern-.125emX}}
\begin{document}

\title{
Explainable Adversarial Attacks on Coarse-to-Fine Classifiers\\
\thanks{This work was supported by NSF under Awards CCF-2106339 and DMS-2304489, and DARPA under Agreement No. HR0011-24-9-0427.} 
}

\makeatletter
\newcommand{\linebreakand}{%
  \end{@IEEEauthorhalign}
  \hfill\mbox{}\par
  \mbox{}\hfill\begin{@IEEEauthorhalign}
}

\makeatother

\author{\IEEEauthorblockN{Akram Heidarizadeh}
\IEEEauthorblockA{\textit{Dept. of Electrical and Computer Engineering} \\
\textit{University of Central Florida}\\
Orlando, FL, USA\\
akram.heidarizadeh@ucf.edu}
\and
\IEEEauthorblockN{Connor Hatfield}
\IEEEauthorblockA{\textit{Dept. of Electrical and Computer Engineering} \\
\textit{University of Central Florida}\\
Orlando, FL, USA\\
connorhat123@gmail.com}
\linebreakand
\IEEEauthorblockN{Lorenzo Lazzarotto}
\IEEEauthorblockA{\textit{School of Technology} \\
\textit{Pontifícia Universidade Católica}\\ 
\textit{do Rio Grande do Sul}\\
Porto Alegre, Brazil \\
l.lazzarotto@edu.pucrs.br}
\and
\IEEEauthorblockN{HanQin Cai}
\IEEEauthorblockA{\textit{Dept. of Statistics and Data Science} \\
\textit{Dept. of Computer Science} \\
\textit{University of Central Florida}\\
Orlando, FL, USA \\
hqcai@ucf.edu}
\and 
\IEEEauthorblockN{George Atia}
\IEEEauthorblockA{\textit{Dept. of Electrical and Computer Engineering} \\
\textit{Dept. of Computer Science} \\
\textit{University of Central Florida}\\
Orlando, FL, USA\\
george.atia@ucf.edu}

}
\maketitle
\begin{abstract}
 Traditional adversarial attacks typically aim to alter the predicted labels of input images by generating perturbations that are imperceptible to the human eye. However, these approaches often lack explainability. Moreover, most existing work on adversarial attacks focuses on single-stage classifiers, but multi-stage classifiers are largely unexplored. In this paper, we introduce instance-based adversarial attacks for multi-stage classifiers, leveraging Layer-wise Relevance Propagation (LRP), which assigns relevance scores to pixels based on their influence on classification outcomes. Our approach generates explainable adversarial perturbations by utilizing LRP to identify and target key features critical for both coarse and fine-grained classifications. Unlike conventional attacks, our method not only induces misclassification but also enhances the interpretability of the model's behavior across classification stages, as demonstrated by experimental results. 
\end{abstract}


\section{Introduction}

The remarkable success of neural networks across diverse domains, from image recognition to natural language processing, has led to their widespread adoption in critical applications such as autonomous driving \cite{kozlowski2024image, maghsoumi2023roadsave}, healthcare \cite{alowais2023revolutionizing}, and security systems \cite{silvestri2023machine} . Despite these advancements, deep neural networks (DNNs) remain vulnerable to adversarial attacks, where imperceptible perturbations to input data can lead to incorrect predictions \cite{szegedy2013intriguing}, \cite{moosavi2016deepfool}, \cite{cai2021zeroth}, \cite{cai2022zeroth}. 

 Traditional attack methods, such as the Fast Gradient Sign Method (FGSM) \cite{goodfellow2014explaining} and Projected Gradient Descent (PGD) \cite{madry2017towards}, have been designed to generate small perturbations within an $\ell_1 $ ball that alter the model's prediction while remaining imperceptible to human observers. While these attacks expose the vulnerability of neural networks, they provide limited insight into the reasoning behind the model’s decision changes, thus lacking explainability. This limitation also applies to hierarchical classifiers, which make decisions across multiple stages. While a few works, such as \cite{alkhouri2021adversarial} and \cite{alkhouri2021Nested}, have explored traditional attacks in hierarchical settings, to our knowledge, no research has addressed explainable attacks in this context. 

Explainability and interpretability have gained significant attention in recent years \cite{baniecki2024adversarial}. Methods like Layer-wise Relevance Propagation (LRP) \cite{bach2015pixel,montavon2017explaining}, GradCAM \cite{selvaraju2017grad}, LIME \cite{ribeiro2016should} and SHAP \cite{lundberg2017unified} provide various mechanisms for explanations by identifying the key features or regions of an input that most influence a model's output.

However, research on explainable adversarial attacks remains limited, even for single-stage classifiers, with only a few studies such as \cite{universalLRP}, \cite{ghorbani2019interpretation}  and \cite{dombrowski2019explanations}, which focus on single-stage universal attacks. The exploration of such methods for coarse-to-fine (C2F) classifiers, where understanding the decision-making process across multiple stages is crucial, remains largely unexplored.

In this paper, we study explainable adversarial attacks on C2F classifiers by introducing an approach that leverages LRP to guide the generation of interpretable adversarial perturbations. 
Unlike traditional attacks that solely aim to change the output label, our method targets key features identified at both the coarse and fine classification stages, as highlighted by LRP-generated heatmaps. By focusing on perturbing these critical features, we generate attacks that not only fool the model but also provide insights into the features the model relies on at each stage of the classification process.

We evaluate the effectiveness of our approach through experiments on a C2F classifier architecture, using benchmark datasets like ImageNet \cite{russakovsky2015imagenet}. Our results show that the generated perturbations successfully mislead the models at both stages of classification, while highlighting a key trade-off between explainability and perceptibility. Compared to the other methods, we allow for greater perturbation, while keeping it imperceptible and achieving enhanced explainability.

\section{Background}
\subsection {Coarse-to-Fine Model Formulation} \label{AA}
Consider a pre-trained hierarchical classifier structured in a C2F hierarchy, where an initial coarse-level classifier provides a broad classification, 
which is then further refined by subsequent finer classifiers.

Each data point $x \in X \subseteq \mathbb{R}^N$ is assigned a coarse label $i \in [M]$, where $M$ denotes the total number of coarse labels and $[M] := \{1, 2, \dots, M\}$. Additionally, there is a fine label $ l \in [M_i]$, where $ M_i$ represents the number of fine classes associated with the $i$-th coarse label. Let $C: \mathbb{R}^N \rightarrow [M]$ be the coarse classifier function that assigns $x$ to a coarse class. 
For classifier $C$, we assume the existence of $M$ discriminant functionals,  $C_i(x): \mathbb{R}^N \rightarrow \mathbb{R}$  for  $i \in [M]$, which are used for coarse classification such that
\begin{equation}
\textstyle
C(x) = \argmax_{i \in [M]} C_i(x).
\label{eq1}
\end{equation}

For each coarse label $i \in [M] $, let $F^i$ 
represent the  $i$-th fine classifier function. Similar to the formulation in \eqref{eq1}, we define $F^i_j(x): \mathbb{R}^N \rightarrow \mathbb{R}$ for $j \in [M_i]$ as the discriminant functions used to determine the finer class of coarse label $i$, such that
\begin{equation}
\textstyle
F^i(x) = \argmax_{j \in [M_i]} F^i_j(x).
\label{eq2}
\end{equation}

\subsection{Layer-wise Relevance Propagation}
Layer-wise Relevance Propagation (LRP) is a technique to explain the decisions made by neural networks, determining the contribution of each parts of the input data to the final decision \cite{bach2015pixel}. To interpret the network’s prediction for a specific class $c$, we propagate relevance scores $R$ from the output layer $L$ back to the input layer, using the activations and network weights. For the output layer, relevance is defined by:
\begin{equation}
R^L_i = \delta_{i,c}\:,
\label{lrp2}
\end{equation}
where $\delta_{i,c}$ is the Kronecker delta, which selects the relevance for class \(c\) by setting \(R^L_i = 1\) when \(i = c\) and \(R^L_i = 0\) otherwise. The relevance scores are then propagated back through all layers except the first using the z+ rule \cite{montavon2017explaining}:
\begin{equation}
R^l_i = \sum_j \frac{a^l_i (W^l)_{ij}^+}{\sum_k a^l_k (W^l)_{kj}^+} R^{l+1}_j,
\label{lrp3}
\end{equation}
where $(W^l)^+$ denotes the positive weights of the $l$-th layer and $a^l$ is the activation vector of the $l$-th layer. Finally, the relevance scores at the input layer are calculated using the $z\beta$ rule \cite{montavon2017explaining}: 

\begin{equation}
R^0_i = \sum_j \frac{a^0_i W^0_{ij} - l_i(W^0)^+_{ij} - h_i(W^0)^-_{ij}}{\sum_k (a^0_i W^0_{kj} - l_i(W^0)^+_{kj} - h_i(W^0)^-_{kj})} R^1_j,
\label{lrp4}
\end{equation}
where $l_i$ and $h_i$ are the lower and upper bounds of the input domain, respectively. For simplicity, we henceforth use the notation $LRP_G(x; c)$ to indicate the relevance scores at the input layer of a classifier $G$, for an input image $x$ and label $c$. 

\section{LRP Attack Formulation}

Unlike the methods in \cite{moosavi2016deepfool}, \cite{madry2017towards}, and \cite{wong2019wasserstein} that directly use the gradients of the DNN's outputs and inputs to generate perturbations, the attacks we propose herein target the heatmaps produced by the LRP method. By assuming that the LRP interpretation indicates the DNN's attention, our algorithm is designed to create perturbations by disrupting this attention.
\begin{figure*}[htbp]
\centerline{\includegraphics[width=0.78\textwidth]{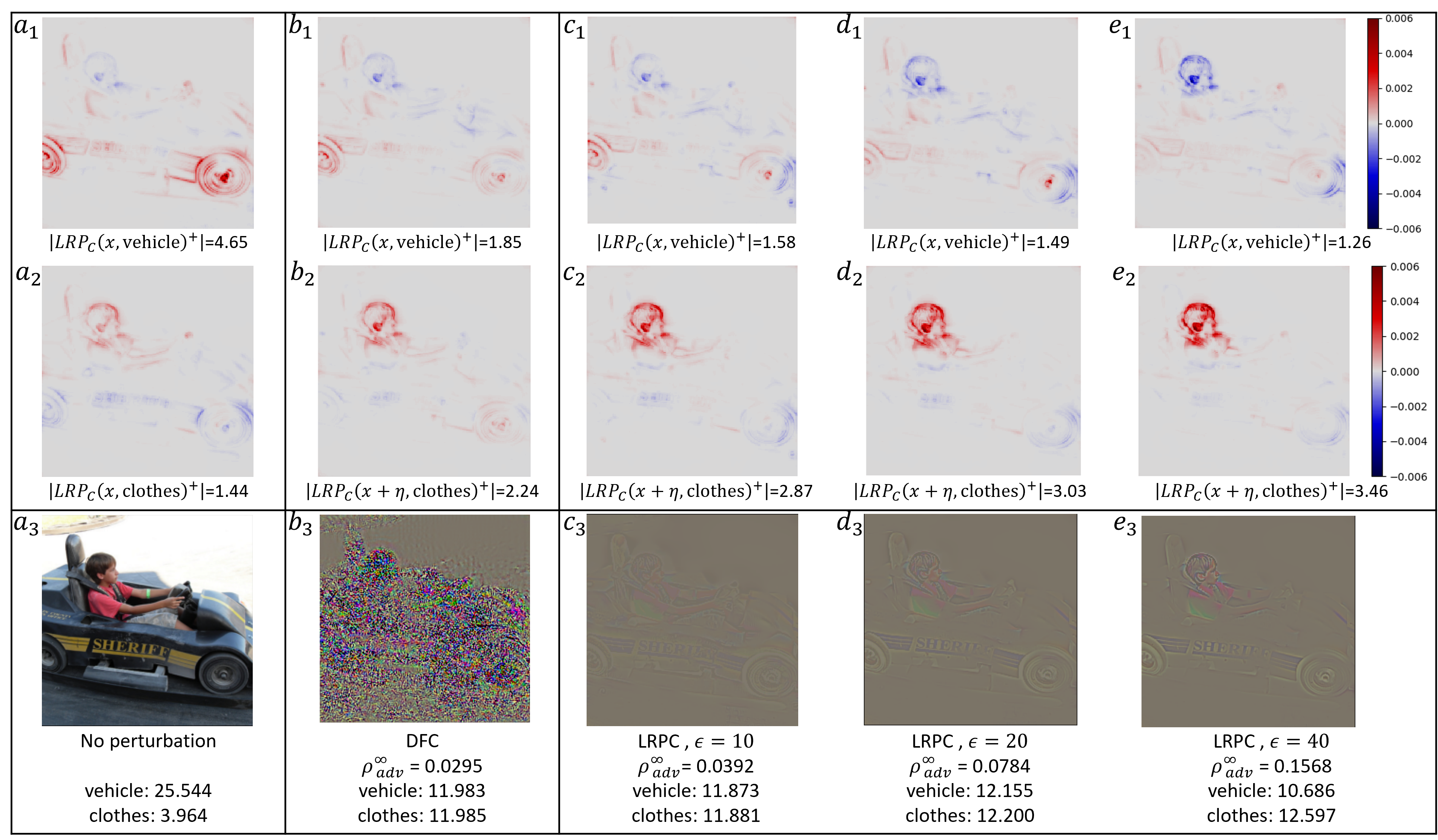}}
\vspace{-.2cm}
\caption{LRP visualizations before and after LRPC and DFC attacks.
(a\textsubscript{1}) LRP of the original coarse class ($r_{\text{org}}$: ``vehicle") before the attack.
(a\textsubscript{2}) LRP of the adversarial coarse class ($r_{\text{adv}}$: ``clothes") before the attack.
(a\textsubscript{3}) Benign image.
(c\textsubscript{1}, d\textsubscript{1}, e\textsubscript{1}) LRP of $r_{\text{org}}$ after LRPC attack for $\epsilon = 10, 20, 40$, compared to (b\textsubscript{1}) for DFC.
(c\textsubscript{2}, d\textsubscript{2}, e\textsubscript{2}) LRP of $r_{\text{adv}}$ after LRPC attack for $\epsilon = 10, 20, 40$, compared to (b\textsubscript{2}) for DFC.
Perturbations generated with LRPC ($\epsilon = 10, 20, 40$) are shown in (c\textsubscript{3}, d\textsubscript{3}, e\textsubscript{3}), and for DFC in (b\textsubscript{3}). LRP norms and prediction scores are displayed below the respective cases.}
\label{fig1}
\end{figure*}

\subsection {Fooling the Coarse Level}
In this attack, the goal is to generate imperceptible additive perturbations to fool the coarse-level classification, satisfying the requirement $C(x +\eta) \neq C(x)$. To formalize this, let $r_{\text{org}} = C(x)$ and $r_{\text{adv}} = C(x+\eta)$ represent the original and adversarial coarse labels, respectively, where these labels correspond to
%
the coarse categories with the highest prediction probability for the input image $x$ and
the perturbed image $x+\eta$. The attacker selects $r_\text{adv}$ as
\begin{align}
\textstyle
r_{\text{adv}} = \argmax_{i \in [M] \setminus r_{\text{org}}} C_i(x)\:,
\end{align}
the coarse label with the second-highest probability after $r_\text{org}$.

To achieve the objective of redirecting the coarse classifier's attention from $r_{\text{org}}$ to $r_{\text{adv}}$, we define a loss function based on the $\ell_p $-norm of the positive and negative relevance scores produced by LRP with respect to these labels. The loss function is designed to decrease all positive relevance scores and increase all negative relevance scores of the heatmap $LRP_C(x + \eta; r_{\text{org}})$, while simultaneously increasing all positive relevance scores and decreasing all negative relevance scores of $LRP_C(x + \eta; r_{\text{adv}})$. 
Thus, we define the loss function for the LRP Coarse-Level Attack (LRPC):
\begin{align}
&\hspace{-.25cm}\mathcal{L}_C = \|LRP_C(x + \eta; r_{\text{org}})^+\|_p - \|LRP_C(x + \eta; r_{\text{adv}})^+\|_p \cr
&- \|LRP_C(x + \eta; r_{\text{org}})^-\|_p + \|LRP_C(x + \eta; r_{\text{adv}})^-\|_p\:.
\label{coarse_loss}
\end{align}
Here, we set $p = 1$, calculating the $\ell_1$-norm of the heatmaps. 
\begin{figure}
\centerline{\includegraphics[width=0.45\textwidth]{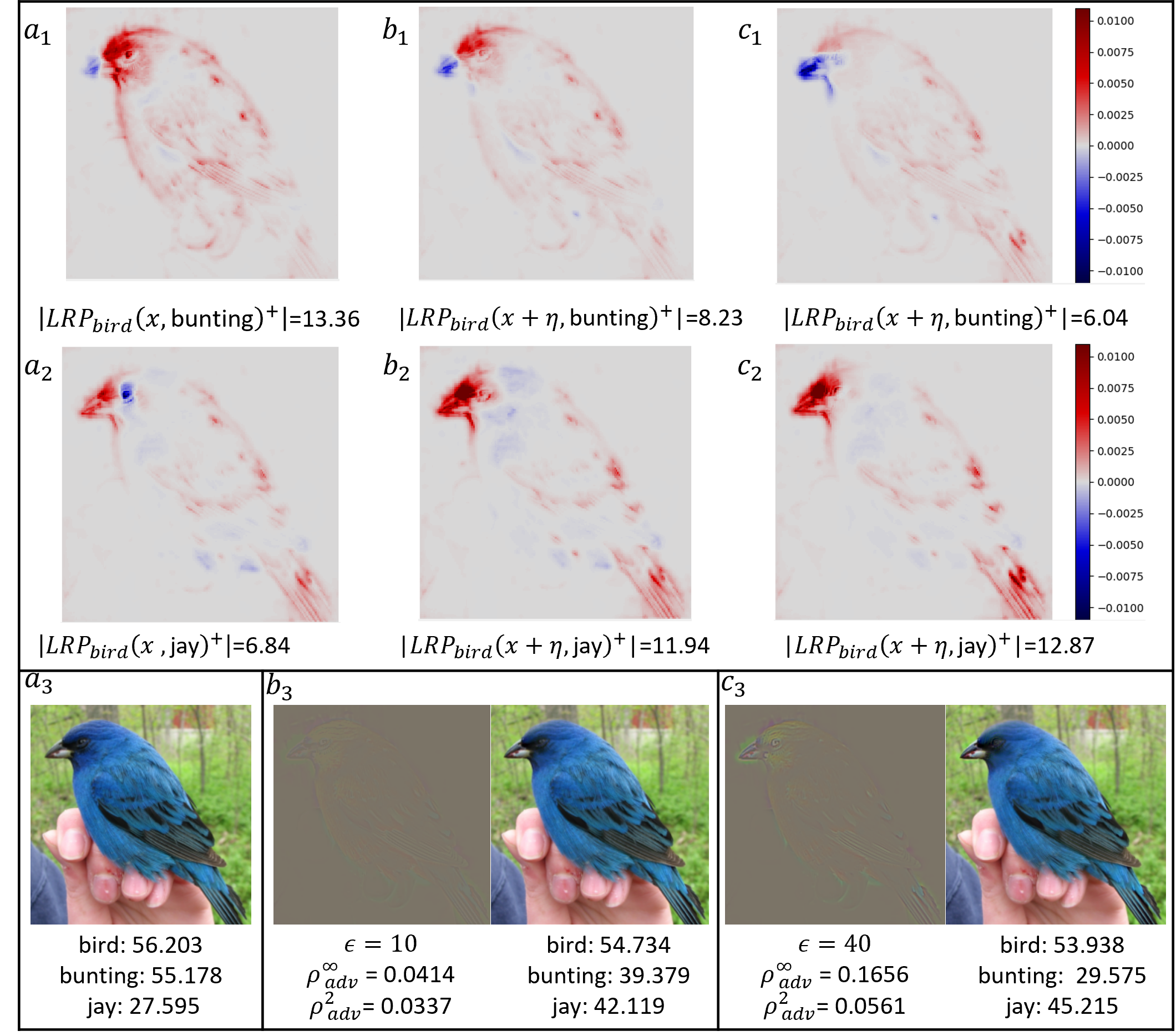}}
\caption{Example of LRP visualization of a\textsubscript{1}) the original fine class ($f_{\text{org}}$) and a\textsubscript2) the adversarial fine class ($f_{\text{adv}}$) before attack. a\textsubscript{3}) Benign image. 
b\textsubscript{1}, c\textsubscript{1}) LRP of $f_{\text{org}}$ after LRPF attack with $\epsilon=10, 40$. 
b\textsubscript{2}, c\textsubscript{2}) LRP of $f_{\text{adv}}$ after LRPF attack with $\epsilon=10, 40$. LRP norms for each case are displayed below the respective heatmaps. b\textsubscript{3}, c\textsubscript{3}) Perturbation with LRPF for $\epsilon=10, 40$ and perturbed image. The prediction scores for each case are displayed below the respective images. $r_{\text{org}}$: ``bird", $f_{\text{org}}$: ``bunting" and $f_{\text{adv}}$: ``jay".}
\label{fig2}
\end{figure}
\subsection {Fooling the Fine Level}
In this scenario, the goal is to create perturbations that alter the finer classification while keeping the coarse prediction unchanged. Specifically, we aim to ensure that $C(x +\eta) = C(x) = r_\text{org}$, while $F^{r_\text{org}}(x)\ne F^{r_\text{org}}(x+\eta)$. The coarse label is first determined and used as a prerequisite to identify the corresponding fine label.
Similar to the coarse level attack, the attacker selects $f_{\text{adv}}$ as:
\begin{equation}
\textstyle
f_{\text{adv}} = \argmax_{j \in [M_{r_\text{org}}] \setminus f_{\text{org}}} F^{r_{\text{org}}}_j(x)\:,
\label{fine_label}
\end{equation}
where $f_{\text{org}} := F^{r_\text{org}}(x)$ and $f_{\text{adv}}$ is the fine label with the second-highest probability. 

We apply the same approach at the fine level by defining the loss function similarly, based on the $\ell_1 $-norm of the positive relevances produced by LRP for the $r_{\text{org}}$-th fine classifier, $F^{\text{org}}$, with respect to the fine-level labels. 
Thus, the loss function for the LRP Fine-Level Attack (LRPF) is defined as: 
\begin{align}
&\mathcal{L}_F\hspace{-.5mm} =\hspace{-.5mm}\|LRP_{F^{r_{\text{org}}}}\hspace{-.5mm}(x + \eta; f_{\text{org}})^+\|_p \hspace{-.5mm}-\hspace{-.5mm} \|LRP_{F^{r_{\text{org}}}}\hspace{-.5mm}(x + \eta; f_{\text{adv}})^+\|_p  \cr
&-\|LRP_{F^{r_{\text{org}}}}(x + \eta; f_{\text{org}})^-\|_p + \|LRP_{F^{r_{\text{org}}}}(x + \eta; f_{\text{adv}})^-\|_p.
\label{fine_loss}
\end{align}
%
Algorithm 1 describes our approach. 
The perturbation is initialized to zero at the beginning of the attack. At each iteration, the perturbation $\eta$ is clipped by $\min(\epsilon, \eta)$, where $\epsilon$ is the maximum permissible $l_\infty$-norm. $\eta$ is then added to the benign images to create perturbed images, which are clipped again to maintain pixel values within the range $[0, 255]$. Both benign and perturbed images are fed into the pre-trained model to obtain the original and adversarial labels. The adversarial labels could change with each iteration, as the goal is to generate small perturbations that progressively move the perturbed image toward the nearest labels at each step. The gradient-descent algorithm is used to minimize the loss function as defined in \eqref{coarse_loss}  and \eqref{fine_loss} and optimize the perturbation $\eta$ via the loss gradients $\frac{\partial \mathcal{L}_C}{\partial \eta}$ and $\frac{\partial \mathcal{L}_F}{\partial \eta}$ for LRPC and LRPF, respectively.
Depending on the attack type, the optimization process terminates either when the coarse label changes (LRPC) or when the fine label changes while maintaining coarse label consistency (LRPF), ensuring effective perturbation generation.
\begin{algorithm}
\caption{LRP-based Attack for Coarse-to-Fine Classifiers}
\begin{algorithmic}[1]
\STATE \textbf{Input:} Pre-trained coarse model $C$ and fine models $F_i$, input image $x$, ground truth labels $(y,z)$, max iterations $K$, step size $lr$, clip parameter $\epsilon$, attack$=$\{LRPC, LRPF\}

\STATE Initialize perturbation $\eta \leftarrow \textbf{0}$, $ k \leftarrow 0$

\STATE Compute labels $r_{\text{org}}$, $r_{\text{adv}}$, $f_{\text{org}}$ and $f_{\text{adv}}$

\WHILE{$k<K$}
    \STATE Calculate LRPs for both labels
    \STATE Compute Loss $\mathcal{L}$ according to \eqref{coarse_loss} or \eqref{fine_loss}
    \STATE Update perturbation $\eta \leftarrow \eta - lr \cdot \frac{\partial \mathcal{L}}{\partial \eta}$
    \STATE $\eta\leftarrow \text{clip}(v; -{\epsilon},{\epsilon})$
    \STATE Update $\mathrm{org}$ and $\mathrm{adv}$ labels for $x+\eta$
    \STATE  $ k \leftarrow k+1$
    \IF{attack$==$LRPC}
        \IF{$r_{\text{org}} \neq y$} 
            \STATE Increment fooling count and break loop
        \ENDIF
    \ELSIF{attack$==$LRPF}
        \IF{$f_{\text{org}} \neq z$ \AND $r_{\text{org}} = y$}
            \STATE Increment fooling count and break loop
        \ENDIF
    \ENDIF        
\ENDWHILE
\STATE \textbf{Output:} perturbation $\eta$
\end{algorithmic}
\end{algorithm}

\section{Experimental Results}
\label{sec:exp}
\vspace{-.1cm}
We evaluate our attack on the C2F architecture using ImageNet \cite{russakovsky2015imagenet} and compare it to other attacks, such as PGD \cite{madry2017towards} and DeepFool \cite{moosavi2016deepfool}. Since these two methods were designed for single-stage classifiers, we adapted them for the C2F setting by applying separate attacks at both the coarse and fine levels. We refer to these adapted versions as DFC and DFF for DeepFool's coarse and fine attacks, and PGDC and PGDF for PGD's coarse and fine attacks, where perturbations in DeepFool are measured using the $\ell_2$-norm, and PGD is implemented with $p=\infty$. 
We calculate the average perceptibility of the attack as:
    \begin{equation}
    \rho^p_{\text{adv}}(f) = \frac{1}{|D|} \sum_{x \in D} \frac{\| \eta \|_p}{\| x \|_p}\:,
    \label{percep}
    \end{equation}
where \( D \) is the dataset, and \( \eta \) is the adversarial perturbation corresponding to input \( x \). 
Additionally, we analyze the fooling ratio, defined as the proportion of images whose labels are changed by the attack relative to the total number of images.
\smallbreak
\noindent\textbf{Coarse-to-fine classification framework.} We have introduced a C2F classifier with a total of $M = 8$ coarse labels for the hierarchical classification of the ImageNet dataset. The classification occurs in two stages: 
a coarse classifier $C$ assigns input images to one of eight broad categories: \{fish, bird, reptile, clothes, food, vehicle, electrical device, dog\}, which are further classified 
by separate fine-level classifiers $F^i$ within each coarse category into fine-grained labels.
\smallbreak
\noindent\textbf{Dataset.} Our dataset is sourced from the ImageNet dataset, which contains 1000 distinct labels. However, only 393 of these labels were used in our model, distributed across the 8 coarse categories. We randomly selected 80\% of the training set from the 393 classes of the ILSVRC2012 train dataset for training the classifiers, with the remaining 20\% used for validation. We evaluate our attack on the VGG-16 network \cite{simonyan2014very}, which has an accuracy of 96\% for the coarse classifier and 78\%-90\% for each of the fine classifiers.
\smallbreak
\noindent\textbf{Explainability-perceptibility tradeoff.} 
Fig.~\ref{fig1} illustrates LRP results before and after the coarse attack for LRPC with varying perturbation levels. The heatmaps highlight regions contributing to coarse-level predictions, showing how the LRPC attack shifts the model's attention from the car to the person inside, which leads to misclassification from ``vehicle" to ``clothes." Both positive and negative relevance scores for the original and adversarial classes are manipulated, with reduced relevance for car pixels (c\textsubscript{1}, d\textsubscript{1}, e\textsubscript{1}) and increased relevance for the person (c\textsubscript{2}, d\textsubscript{2}, e\textsubscript{2}). As perturbation increases (via higher $\epsilon$), explainability improves, revealing a tradeoff between explainability and perceptibility. Our results clarify this relationship, both quantitatively (by illustrating the tradeoff between perceptibility and relevance scores), and qualitatively (by progressively emphasizing key regions while reducing attention on others as perturbation increases). In contrast, the DFC attack (b\textsubscript{1}, b\textsubscript{2}, b\textsubscript{3}) scatters the perturbation, whereas our attack concentrates it on critical image regions. Fig.~\ref{fig2} demonstrates that our algorithm is also explainable at the finer level, although the features are very similar in this stage. In this example, the attack shifts the model's attention from the eye and body of the ``indigo" to areas typically associated with a ``jay", such as the beak and tail feathers, leading to misclassification.
\smallbreak
\noindent\textbf{Performance.}
We evaluate the performance of our attack for both coarse- and fine-level attack algorithms. Table~\ref{tab:attack_comparison1} shows $\rho_{\text{adv}}^1$, $\rho_{\text{adv}}^2$, $\rho_{\text{adv}}^\infty$ and fooling ratio for LRPC, DFC, and PGDC, tested on the first $1000$ RGB images of the validation set.  
LRPC achieves perceptibility comparable to PGDC in both $\ell_1$ and $\ell_\infty$-norms. Moreover, for $\epsilon = 10$, the $\ell_\infty$-norm is on par with DFC, while the fooling rate remains high. 
Table~\ref{tab:attack_comparison2} provides the fooling ratio and perceptibility values for the fine-level attack algorithms. LRPF achieves high fooling rates even with smaller perturbations, indicated by lower $\epsilon$ values. This demonstrates that LRPF can maintain competitive fooling rates while effectively controlling perceptibility. We emphasize that our approach prioritizes explainability rather than asserting quantitative superiority in metrics like fooling ratio or perceptibility.
\section {Conclusion}
We developed an approach for explainable adversarial attacks on coarse-to-fine classifiers by leveraging Layer-wise Relevance Propagation (LRP) to generate interpretable perturbations. Our method targets critical features identified at both classification stages, providing insights into the model's decision-making process while successfully misleading the model and outperform traditional methods in providing clearer interpretations without compromising attack imperceptibility.

\begin{table}[th]
\vspace{-.3cm}
\caption{Fooling ratio and perceptibility of coarse-level attacks.}
\vspace{-.2cm}
\centering
\begin{tabular}{c c c c c c }
\toprule
\textbf{Algorithm} & \makecell{\textbf{LRPC} \\ $\epsilon = 10$} & \makecell{\textbf{LRPC} \\ $\epsilon = 20$} & \makecell{\textbf{LRPC} \\ $\epsilon = 40$} & \textbf{DFC} &  \textbf{PGDC}  \\
\midrule
\vspace{0.03in}
$\rho_{\text{adv}}^2$ &  0.0294 & 0.0323 & 0.0405  & 0.0045 & 0.0262
 \\
\vspace{0.03in}
$\rho_{\text{adv}}^1$ & 0.0216  & 0.0174 &  0.0195 & 0.0031 & 0.0224
 \\
\vspace{0.03in}
$\rho_{\text{adv}}^\infty$&  0.0399 & 0.0778& 0.1557 & 0.0408
 & 0.0101 \\
 \vspace{0.03in}
Fooling(\%) & 87.1 & 92.5 & 99.3 & 100 & 100 \\
\bottomrule
\end{tabular}
\label{tab:attack_comparison1}
\end{table}

\begin{table}[th]
\vspace{-.3cm}
\caption{Fooling ratio and perceptibility of fine-level attacks.}
\vspace{-.2cm}
\centering
\begin{tabular}{c c c c c c}
\toprule
\textbf{Algorithm} & \makecell{\textbf{LRPF} \\ $\epsilon = 10$} & \makecell{\textbf{LRPF} \\ $\epsilon = 20$} & \makecell{\textbf{LRPF} \\ $\epsilon = 40$} & \textbf{DFF} & \textbf{PGDF} \\
\midrule
\vspace{0.03in}
$\rho_{\text{adv}}^2$ & 0.0127  & 0.0145 & 0.0151 & 0.0020 & 0.0078 \\
\vspace{0.03in}
$\rho_{\text{adv}}^1$ &   0.0084  & 0.0079 & 0.0066 & 0.0013 & 0.0092 \\
\vspace{0.03in}
$\rho_{\text{adv}}^\infty$ &  0.0241 & 0.0542 & 0.0819 & 0.0029 & 0.0035 \\
\vspace{0.03in}
Fooling(\%) & 98.7 & 100 & 100 & 100 & 95.7 \\
\bottomrule
\end{tabular}
\label{tab:attack_comparison2}
\end{table}

\newpage
\bibliographystyle{IEEEtran}

\vspace{12pt}

\end{document}